\documentclass[letterpaper, 10 pt, conference]{ieeeconf}

\IEEEoverridecommandlockouts                               % This command is only needed if 
\usepackage{graphics} % for pdf, bitmapped graphics files
\usepackage{epsfig} % for postscript graphics files
\usepackage{mathptmx} % assumes new font selection scheme installed
\usepackage{times} % assumes new font selection scheme installed
\usepackage{amsmath} % assumes amsmath package installed
\usepackage{amssymb}  % assumes amsmath package installed
\usepackage{graphicx}
\usepackage{multirow}
\usepackage{tabularx,booktabs}
\usepackage{hyperref}
\usepackage{cite}
\usepackage{fancyhdr}
\fancypagestyle{firstpage}{
    \chead{This paper has been accepted for publication at the 2026 IEEE/RSJ International Conference on Intelligent Robots and Systems (IROS 2026)}}
\usepackage{comment}
\usepackage{booktabs}
\usepackage[switch]{lineno}
\usepackage{array}
\usepackage{xcolor}
\usepackage{soul}
\usepackage{censor}
\usepackage{siunitx}
\usepackage{algorithm}
\usepackage{makecell}
\usepackage{algpseudocode}
\usepackage{float}
\usepackage{graphics} % for pdf, bitmapped graphics files
\usepackage{epsfig} % for postscript graphics files
\usepackage{mathptmx} % assumes new font selection scheme installed
\usepackage{times} % assumes new font selection scheme installed
\usepackage{amsmath} % assumes amsmath package installed
\usepackage{amssymb}  % assumes amsmath package installed
\usepackage{graphicx}
\usepackage{multirow}
\usepackage{tabularx,booktabs}
\usepackage{hyperref}
\usepackage{cite}
\usepackage{comment}
\usepackage{booktabs}
\usepackage[switch]{lineno}
\usepackage{array}
\usepackage{xcolor}
\usepackage{soul}
\usepackage{siunitx}
\usepackage{pgfplots}

\usepackage{float}

\title{\LARGE \bf
A Physics-Based Closed-Loop Robotic Bioprinting \\ Framework Towards Volumetric Muscle Loss Treatment}

\author{Omid Rezayof$^{1}$, Jerin T. Andrews$^{2}$, Ehsan Zobeidi$^{1}$, Ali Ghasemkhani$^{1}$, Meenakshi Kamaraj$^{3}$, \\Maryam Tilton$^{1}$, Johnson V. John$^{3}$, Farshid Alambeigi$^{1}$ % <-this % stops a space
\thanks{© 2026 IEEE. Personal use of this material is permitted. Permission from IEEE must be obtained for all other uses, in any current or future media, including reprinting/republishing this material for advertising or promotional purposes, creating new collective works, for resale or redistribution to servers or lists, or reuse of any copyrighted component of this work in other works.}
\thanks{Research reported in this
publication was supported by the National Institute Of Arthritis And Musculoskeletal And Skin Diseases of
the National Institutes of Health under Award Numbers DP2AR082471.}
\thanks{$^{1}$O.~Rezayof, E. Zobeidi, A. Ghasemkhani, M. Tilton, and F.~Alambeigi are with the Walker Department of Mechanical Engineering, University of Texas at Austin, TX, USA.
Email: {\tt\small \{omid.rezayof, ehsan.zobeidi, ali.ghasemkhani\}@utexas.edu}, and {\tt\small \{maryam.tilton, farshid.alambeigi\}@austin.utexas.edu}}.
\thanks{$^{2}$ J. T. Andrews is with Chandra Department of Electrical and Computer Engineering, University of Texas at Austin, TX, USA. Email:{\tt\small jerin.andrews@utexas.edu}}  
\thanks{$^{3}$ M. Kamaraj and J. V. John are with Terasaki Institute for Biomedical Innovation, Los Angeles, CA, USA. 
Email:{\tt\small \{meenu.kamaraj, jjohn\}@terasaki.org}}}

\begin{document}

\maketitle
\thispagestyle{firstpage}
\pagestyle{empty}

%%%%%%%%%%%%%%%%%%%%%%%%%%%%%%%%%%%%%%%%%%%%%%%%%%%%%%%%%%%%%%%%%%%%%%%%%%%%%%%%
\begin{abstract}
Robotic bioprinting and Direct Ink Writing (DIW) are being explored towards the treatment of Volumetric Muscle Loss (VML). While previous studies have shown the importance of proper parameter selection on the print outcome, existing approaches often rely on time- and material-intensive design of experiments methods, or require large, well-curated datasets for training machine learning models.
In this paper, we propose a physics-based closed-loop robotic bioprinting system capable of near real-time parameter adaptation. The system integrates a 3D point cloud camera and fully autonomous vision-based algorithms to provide quantitative evaluation of printed constructs. This evaluation is fed into a controller that adjusts printing parameters to achieve a desired bead thickness.
To assess the framework's performance, four experimental configurations were tested, each repeated three times. In these tests, printing began from an arbitrary initial parameter value, and the controller was tasked with adjusting the parameters to reach the desired thickness. The system converged in all trials, achieving a tracking error below 0.5 mm within an average of 5.2 seconds from the start of printing. The low standard deviation of the converged pressure over different tests (0.04 bar on average) demonstrates robustness and repeatability. Additional experiments were conducted with the controller turned off, enabling direct comparison with open-loop DIW bioprinting, further confirming the effectiveness of the proposed closed-loop framework in achieving the desired bead geometry.
\end{abstract}

%%%%%%%%%%%%%%%%%%%%%%%%%%%%%%%%%%%%%%%%%%%%%%%%%%%%%%%%%%%%%%%%%%%%%%%%%%%%%%%%
\section{INTRODUCTION}\label{sec:intro}
Extensive skeletal muscle injuries, such as those resulting from Volumetric Muscle Loss (VML), often surpass the tissue's natural regenerative capacity and require clinical intervention to restore muscle functionality and prevent long-term deficits \cite{Jrvinen2007MuscleIO}. 
Conventional treatments such as autologous tissue transfer, often lead to inadequate tissue regeneration due to implant-tissue mismatch and complications at the injury site \cite{Grogan2011VolumetricML}. 
Recently, bioprinting is being explored to overcome existing limitations in VML injury treatment \cite{Jana2016AnisotropicMF}.   
While several handheld bioprinters are proposed in the literature \cite{OConnell2016DevelopmentOT, Rezayof2024OnTP}, robotic bioprinting is transforming the field of tissue engineering, by offering precise, efficient, and rapid fabrication of tissue constructs \cite{Murphy20143DBO}.
Among the various bioprinting methods, material extrusion and specifically Direct Ink Writing (DIW) has become the most widely used approach for depositing bioinks, across a broad range of applications.

\begin{figure*}[]
   \centering
   \includegraphics[width=0.85\linewidth]{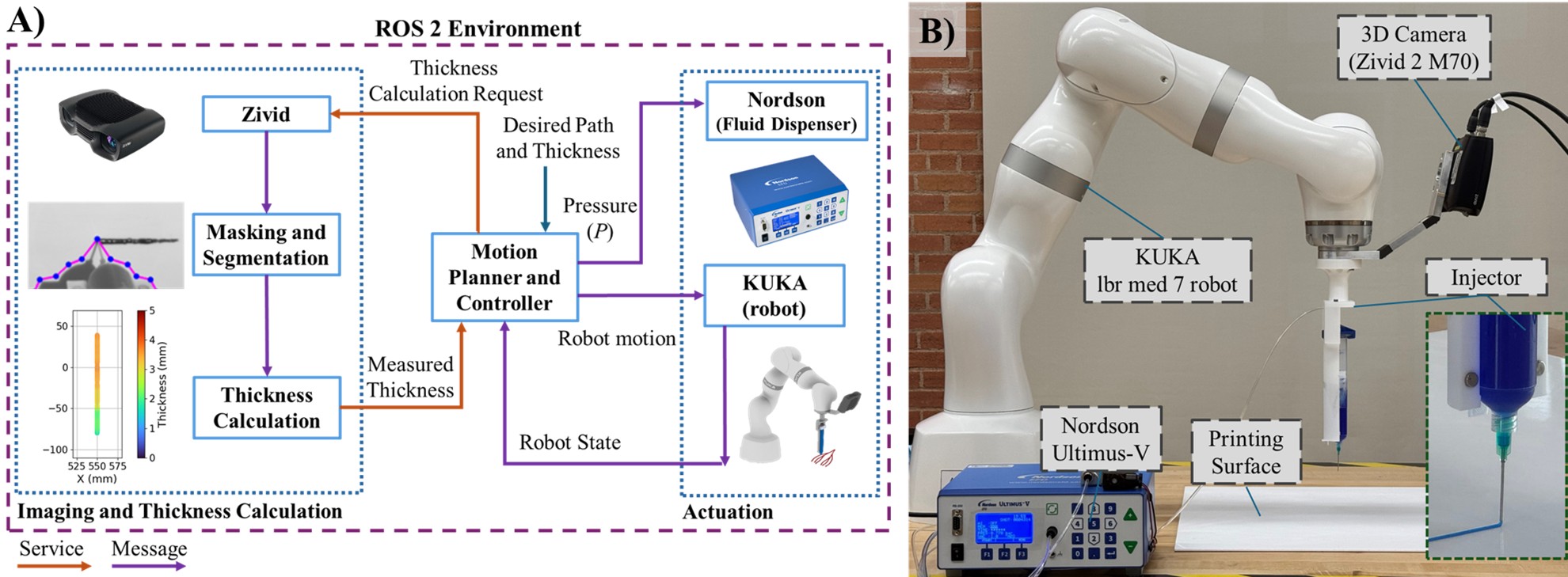}
   \caption{A) The system architecture of the closed-loop printing framework. The figure shows the different modules withing the framework including the main motion planner and controller, the imaging and thickness calculation module, and the actuation module consisting of the robot and the pneumatic injector. B) The figure shows the actual experimental setup including the robot, the 3D camera and the injection system. Furthermore, common terminology used in the paper including the bead thickness ($d$), height (needle height, $H$), and robot velocity ($v_r$) are shown. }
   \label{fig:framework}
   \vspace{-4mm}
\end{figure*}

The outcome of a bioprinting task is influenced by the choice of printing parameters, such as injection rate, nozzle-to-surface distance (i.e., needle height) and nozzle transverse velocity.
Despite the efforts in the literature for intelligent selection of printing parameters to achieve consistent and high-fidelity prints, these parameters are still usually tuned using demanding offline experiments. These methods, commonly referred to as Design-of-Experiments (DoE) methods, involve iterative testing to find optimal parameters for a certain printing task and a specific printing material\cite{AbdulHaq20193DPP, Propst2025TimeCF, multiLIROS26}.
Nevertheless, these methods are extremely time and material consuming, and have been shown to become impractical when the biomaterial's properties change over time \cite{Liu2019ImageAC}. These challenges arise from the open-loop structure of these workflows, which solely rely on fixed preset printing parameters and do not reflect the current state during the printing procedure.
To address these issues, model-based control strategies have been proposed that leverage mechanical and fluid dynamical modeling to predict and control the deposition rate in DIW settings \cite{Hildner2021, Kim2019OptimizationO3}. 
For instance, Hildner used classical fluid mechanics formulae to model different material behavior such as pressure drop and friction to model the deposition rate in bioprinting applications \cite{Hildner2021}.  
While these methods have sh
own promising results in controlling the deposition rate in simulation, they require precise material characterization and usually rely on simplifying assumptions which limit their scalability in real-world applications. Notably, printable biomaterials typically encompass behaviors such as non-Newtonian or time-variant properties. These complex behaviors add significant challenges to the development of accurate predictive models and the implementation of material-specific model-based control strategies \cite{Propst2025TimeCF, Kamaraj2024DevelopmentOS}.

Recently, researchers are exploring closed-loop bioprinting systems for controlling the printing parameters \cite{Fang2024ProcessMD, Singh2023TowardCA, Kopatz2023PressureBasedPM, Krauss2015ThermographicPM, Liu2019ImageAC}. These systems consider output feedback to enable near real-time printing parameters adjusting. 
Several recent studies have shown promising results using Machine Learning (ML) and data-driven approaches such as computer vision, Neural Networks and even Reinforcement Learning \cite{Bonatti2022ADL, Kelly2025, Roach2023InvertibleNN, Piovari2022ClosedloopCO}. However, these ML-driven solutions require large and well-structured datasets and still lack generalizability across different inks when trained on some specific material samples. 
Furthermore, typically, ML-based methods do not provide direct, physically interpretable measurements of bead thickness, and instead, rely on features from latent spaces or hidden layers, which lack clear correspondence to measurable quantities \cite{Bonatti2022ADL, Kelly2025}. 
As a result, these systems often implement binary or heuristic control strategies (e.g., on/off control) without access to the actual thickness values being deposited \cite{Bonatti2022ADL, Kelly2025}. 
Meanwhile, although some efforts have been made to autonomously measure bead thickness as a physical value in the literature \cite{Yang2024DevelopmentAQ, Rezayof2024QuantitativeEO, Roach2023InvertibleNN}, these methods often rely on user inputs, hence lacking the ability to operate in a fully autonomous and near real-time manner.

To overcome the above-mentioned limitations, in this study and as  our main \textit{contribution}, we present a novel physics-based closed-loop robotic bioprinting framework for achieving desired thicknesses of the printing beads. Without relying on pre-trained models, large datasets or material-specific formulations, our framework continuously monitors and evaluates the printing output and adaptively adjusts the printing parameters in near real time to maintain a desired performance, even when printing materials with complex behaviors. 
The proposed robotic bioprinting framework uses a structured-light 3D camera along with custom-developed computer vision algorithms to measure the deposited filament's thickness, and utilizes a controller to dynamically adjust the printing parameters, namely the printing pressure and the needle height, during the print. 
The framework has been rigorously evaluated across various printing conditions, validating the effectiveness of the proposed framework.

\section{Methodology}

\subsection{Proposed Framework and Problem Definition}
Here, we present a framework for near real-time measurement and printing parameters control in DIW bioprinting.
The proposed framework takes a substantial step forward toward adaptive ink writing and bioprinting. 
It replaces typical DoE workflows and eliminates the need for training datasets or material-specific fluid dynamics modeling by leveraging vision-based feedback control instead.
This is particularly beneficial when printing biomaterials which exhibit complex behaviors such as non-Newtonian properties or time-dependent variability as developing accurate fluid dynamical modeling for such materials is not trivial. Furthermore, in conventional DIW frameworks, the internal fluid dynamics within the syringe, or factors like the remaining material volume, are often neglected in the parameter tuning process. However, incorporating real-time feedback and adaptive control can effectively compensate for these effects during the printing process. 

The framework's core concept is shown in Fig. \ref{fig:framework}A. The goal of the framework's control loop is to guide the thickness of the printed beads to track a reference signal (i.e., a desired thickness). This is achieved using a closed-loop feedback  control loop, where a sensing module (i.e., imaging and thickness calculation) is measuring the bead's thickness and a controller module dynamically adjusts the printing parameters according to the difference between the desired and actual thicknesses.  
Of note, in this work, we control the printing pressure $P$ and the needle height $H$, while other printing parameters are constant across the prints.
While the printing pressure is controlled directly as a control input, the needle height is controlled indirectly as a function of the printing pressure ($H(P)$) using a reduced order physics-based model. The details of the control logic can be found in Sections~\ref{subsec:p_controller} and \ref{subsec:h_controller}. 

Fig.~\ref{fig:framework}A further shows the detailed system architecture of the framework, highlighting its different modules. After designing a printing path and ideal thickness, the print starts with an initial pressure value. 
Depending on the use, this pressure can be an arbitrary value (if no prior information is known about an ideal pressure) or be set to a pre-determined pressure. The framework will adapt the printing pressure based on the measured bead thickness to mitigate any effects imposed by the material change or the system dynamics during the print. The details of the thickness calculation and the control modules can be found in later sections. 
\begin{figure}[]
   \centering
   \includegraphics[width=0.85\linewidth]{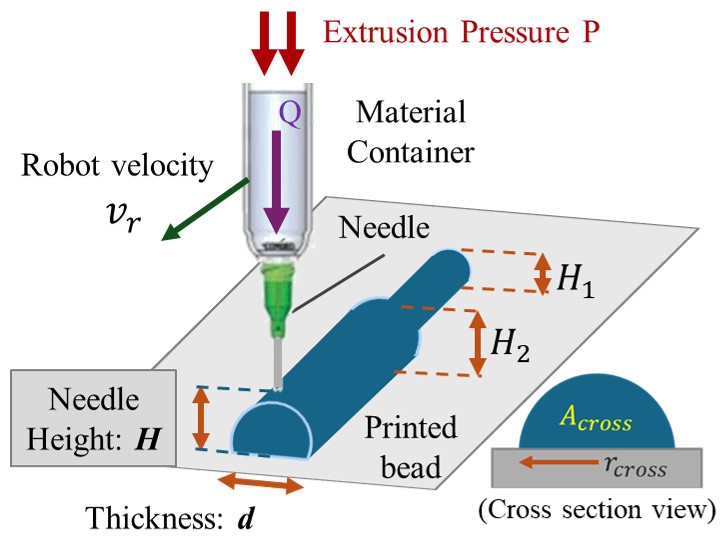}
   \caption{Figure shows the common terminology used in the paper, including the printing pressure $P$, deposition (volume) rate $Q$,  robot transverse velocity $v_r$, needle height $H$, bead thickness $d$, and the cross section of the printed bead. }
   \label{fig:term}
   \vspace{-2mm}
\end{figure}

\subsection{Direct Control of Printing Pressure}
\label{subsec:p_controller}
The framework employs an integral controller at the core of its control module for adjusting the printing pressure $P$. 
Specifically, the control input $u$ (here, the injection pressure) is determined by $u=\int K_i.e(t) dt$ where $K_i$ is the integral gain and $e$ is the tracking error (i.e., the difference between the desired and measured thickness).
For discrete systems, this relationship becomes $u = \Sigma K_i.e.\Delta t$ where $\Delta t$ denotes the output sampling interval. If the sampling interval is approximately constant, $\Delta t$ can be set to one, and its effect can be incorporated into the controller gain $K_i$.  
Despite its simple structure, the experimental results indicate that the controller achieves sufficient accuracy and eliminates the steady-state error.

\subsection{Indirect Control of Needle Height}
\label{subsec:h_controller}
Printing pressure is used as an independent control input to adjust the material deposition rate to control the thickness of the beads.
However, changing the material flow rate affects not only the thickness of the beads, but also the height of the deposited material relative to the printing surface (see Fig.~\ref{fig:term}). 
Hence, controlling the needle height is essential to prevent the needle from penetrating the beads.

In general, needle height ($H$) can be controlled independently in the printing process to achieve desired behaviors. 
However, to reduce controller complexity, we consider a \textit{reduced order physics-based} model by taking $H$ as a dependent variable.
A natural choice is to take $H$ as a function of the deposition rate $Q$ and the robot transverse velocity, which is constant in this study ($H = f_1(Q), v_r=cte$). 
The deposition rate $Q$ can further be estimated using fluid mechanics models relating the injection pressure $P$, material viscosity $\mu$ and the material flow characteristics. It is important to note that, the material viscosity $\mu$ may be a function of time $t$ (for time-variant materials) and/or the material shear rate $\dot{\gamma}$ (for non-Newtonian fluids) $\mu = f_2(\dot{\gamma},t)$ \cite{NonNewtonianFluidsIrgens2014}. As bioinks often posses non-Newtonian properties, it is important to choose a model which can effectively reflect these complex features.

Several models have been proposed in the literature for describing the dependence of viscosity on shear rate for non-Newtonian fluids \cite{NonNewtonianFluidsIrgens2014, Sochi2015AnalyticalSF}.
The Carreau fluid model  in (\ref{eq:carreau})  describes the non-Newtonian behavior more accurately than simpler models such as the Power Law \cite{NonNewtonianFluidsIrgens2014, Sochi2015AnalyticalSF}. 
\begin{equation}
    \mu(\dot{\gamma}) = \mu_i+(\mu_0-\mu_i)(1+\lambda^2\dot{\gamma}^2)^{(n-1)/2}
    \label{eq:carreau}
\end{equation}
where $\mu_i$ is the viscosity at infinite shear rate, $\mu_0$ is the zero shear rate viscosity, and $\lambda$ and $n$ are material-specific model properties.
The models can be fitted to experimental data to first identify the model parameters, and then calculate the ideal needle height $H$ as a function of injection pressure $P$ as explained below. 

Assuming non-compressible fluids, the material flow rate can be approximated by the flow rate in the needle under the printing pressure $P$. The flow rate $Q$ may be calculated using developed equations in the literature based on a selected viscosity model (here Carreau model) as detailed below.
To approximate the flow rate in a circular pipe of length $L$, radius $r$, and under pressure $P$, (\ref{eq:QCarreau_1}) was developed in the literature for Carreau fluids \cite{Sochi2015AnalyticalSF}. 
\begin{equation}
    \label{eq:QCarreau_1}
    Q_{Carreau}(P) = \frac{8\pi.L^3I}{P^3}
\end{equation}
where $I$ is the result of an analytical bounded integration as in 
\begin{equation}
    \label{eq:QCarreau_2}
    I = \int_{0}^{\dot{\gamma}_w} f_3(\dot{\gamma};\mu_0, \mu_i, n, \lambda) \,d\gamma
\end{equation}
\begin{figure}[]
   \centering
   \includegraphics[width=1.0\linewidth]{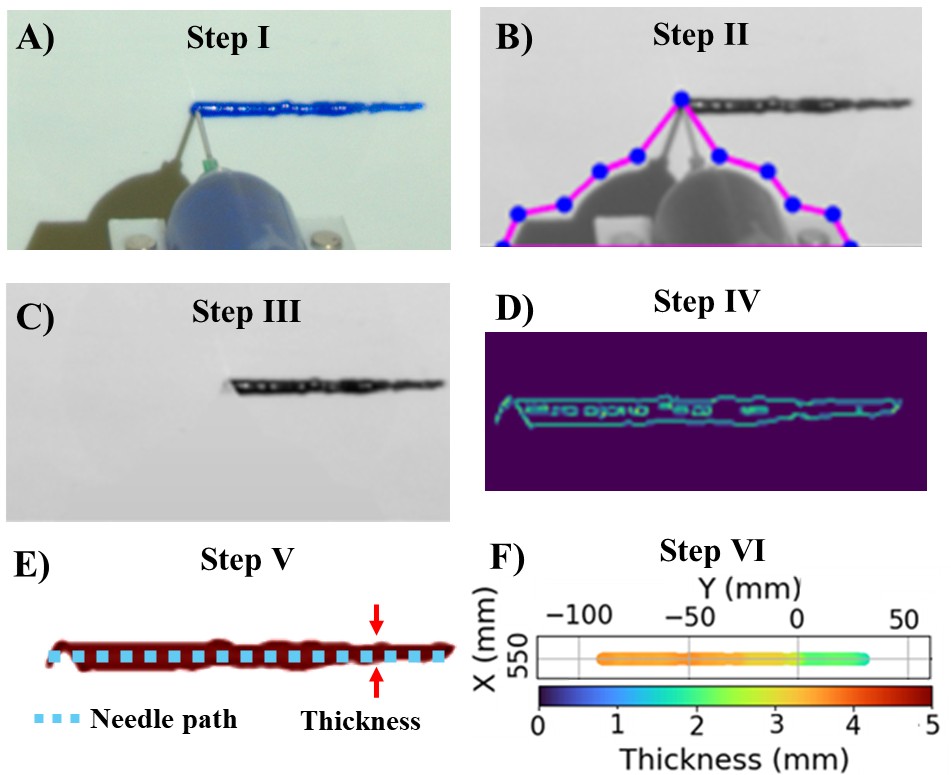}
   \caption{Delineation and thickness calculation process. (A) Step 1: Scanning. RGB image taken using the camera is cropped to a Region of Interest (RoI). 
   (B) Step II: Grayscale conversion and mask definition. The image is converted to grayscale and a closed polygon mask is defined. 
   (C) Step III: Masking. The printer tip is masked using the defined polygon. 
   (D) Step IV: Edge detection. Canny edge detection algorithm is used to identify edges. 
   (E) Step V: Contour tracing. The contours are joined together to form closed loop contours representing the printed beads. The needle path is overlaid on the identified contours.
   (F) Step VI: Thickness calculation. Thickness is calculated for each point in needle path.}
   \label{fig:thickness}
\end{figure}
\noindent where $f_3$ is a function of $\dot{\gamma}$ and known parameters of $\mu_0, \mu_i, n, \lambda$ based on the fitted Carreau model. The detail of the closed form definition of $f_3$ can be found in \cite{Sochi2015AnalyticalSF}. The upper bound of the integration ($\dot{\gamma}_w$, or the shear rate at the wall) can also be solved numerically using (\ref{eq:QCarreau_3}).
\begin{equation}
    \label{eq:QCarreau_3}
    [\mu_i+(\mu_0-\mu_i)(1+\lambda^2.\dot{\gamma}_w^2)^{(n-1)/2}].\dot{\gamma}_w = \frac{rP}{2L}
\end{equation}
\noindent By solving \ref{eq:QCarreau_3} for $\dot{\gamma}_w$, the integral in \ref{eq:QCarreau_2} can be evaluated, and subsequently, the flow rate in \ref{eq:QCarreau_1} can be calculated. 
Having the flow rate $Q$, we can estimate the height of the material which is being laid down on the printing surface based on conservation of mass, and set the needle height to this value. 
\begin{equation}
    \label{eq:conservation}
    Q=v_r.A_{cross}=v_r(\pi r_{cross}^2/2)
\end{equation}
where $A_{cross}$ and $r_{cross}$ are the bead's cross sectional area and radius respectively as shown in Fig.~\ref{fig:term}.  
Assuming the desired needle height $H$ to be equal to the material height on the surface, \ref{eq:H_function} provides the relationship held between the printing pressure $P$ and the desired needle height $H$ given the material flow rate as a function of pressure $Q(P)$. 
\begin{equation}
    \label{eq:H_function}
    H=f_1(Q(P))=\sqrt{\frac{2Q(P)}{v_r\pi}}
\end{equation}

To avoid the complex formulation which is time consuming and not suitable for real-time applications, in this study, we estimate $H=f_1(Q(P))$ as a linear function called the Dynamic Height Adjusting (DHA) function such as $DHA=H(P) = h_0+\alpha.P$ where $\alpha$ and $h_0$ are found based on fitting the model to the experimental data. The values and more detail can be found in Section~\ref{sec:results_and_discussion}. 

\subsection{Robotic Bioprinting Setup} \label{subsec:hardware}
As shown in Fig.~\ref{fig:framework}B, the proposed robotic bioprinting framework includes a seven Degree-of-Freedom (DoF) robotic manipulator (KUKA LBR Med 7, KUKA, Germany), an air-pressure driven custom designed bioprinter (i.e., injector) and a structured-light 3D camera (Zivid 2 M70, Zivid, Norway). 
The Zivid camera is capable of capturing both RGB images and 3D point clouds from the printing site. The resolution of the point cloud scans are specified as 0.2 mm by the manufacturer in a working distance of 300 mm. 
The bioprinter tool, includes a pneumatic fluid dispenser (Nordson Ultimus V, Nordson Corporation, U.S.A.) with a commercial blunt tip Gauge 18 needle as the dispensing nozzle. The syringe (i.e. material container) is held by a custom designed and fabricated fixture, which is attached to the robot's EE. 
Second generation of Robot Operating System (ROS2-Humble) is used as the control and communication environment between the robotic arm, 3D camera, fluid dispenser, pneumatic valve and sensing and control modules.
To command the robot via ROS, Fast Robot Interface (FRI) provided by KUKA was utilized, transmitting commands at a frequency of 100 Hz \cite{Huber2024}. 
The algorithms were executed on a workstation running Ubuntu 22.04, with a 12th Gen Intel Core i9-12900K processor and equipped with an NVIDIA GeForce RTX 3090 Ti GPU, and 64 GB of RAM.

\subsection{Hardware Calibration and Registrations}
Multiple calibrations were performed to achieve accurate hardware integration as the framework integrates multiple individual hardware components. 
Specifically, a Pivot Calibration was performed to identify the location of the bioprinter's tip in robot's workspace, and an Eye-in-Hand Calibration was performed to transform the point cloud data from camera coordinate frame into the robot's frame. 
\subsubsection{Pivot calibration} Pivot calibration was performed to find the relative offset between the End Effector (EE) frame of the robotic arm and the tip of the dispensing needle. Having this offset will allow accurate task space control of the nozzle tip. 
The calibration is performed by recording the pose (position and orientation) of the robot's EE relative to the robot's world frame, while rotating the printer around its needle's tip at a stationary point in space (i.e., the pivot).
The offset ($b_{tip}\in\mathbb{R}^3$) is determined by calculating the location of the stationary pivot point  ($b_{post}\in\mathbb{R}^3$)
for each recorded pose, as in $b_{post} = R_jb_{tip}+p_j$, where $p_j\in\mathbb{R}^3$ and $R_j\in SO(3)$ refer to the EE's position and rotation for each pose $j$ in the total $J$ number of poses. 
Finally, solving the over-determined system of equations in (\ref{eq:pivot_eq_2}) using a least-square approach will yield $b_{tip}$ and $b_{post}$ simultaneously.

\begin{equation}\label{eq:pivot_eq_2}
\begin{bmatrix} 
... & ... \\ 
R_j & -I \\ 
... & ... 
\end{bmatrix}_{(3J\times6)}
\begin{bmatrix} 
b_{tip} \\ 
b_{post} 
\end{bmatrix}_{(6\times1)}
= 
\begin{bmatrix} 
... \\ 
-p_j \\ 
... 
\end{bmatrix}_{(3J\times1)}
\end{equation}

\subsubsection{Eye-in-hand calibration} 

As previously noted, the camera is rigidly mounted on the robot’s EE. To accurately map the camera’s scans into the robot’s coordinate system $\{W\}$, the spatial relationship (i.e. rotation and translation) between the EE frame $\{EE\}$ and the camera frame $\{C\}$ must first be established.
Thus, eye-in-hand calibration is performed to find the missing transformation $X \in SE(3)$ between the camera's frame $\{C\}$ and the robot's EE frame $\{EE\}$.

To perform the calibration, the robot is moved to $M$ number of different poses in the space and the camera is used to take scans from a spatially fixed calibration board. 
At each pose $m$, the camera calculates the transformation between the calibration board and itself ($S_m\in SE(3)$), and the transformation between the robot's world frame to the EE frame, denoted by $E_m\in SE(3)$, is found using the forward kinematics of the robot.
It can be seen that for two poses $m$ and $m'$, the following holds:
\begin{equation}\label{eq:eye_in_hand_1}
E_m.{X}.S_m = E_{m'}.{X}.S_{m'}
\end{equation}
and therefore:
\begin{equation}\label{eq:eye_in_hand_2}
{X}.S_m.S_{m'}^{-1} = E_m^{-1}.E_{m'}.{X}
\end{equation}
By selecting every pair of poses among the total of $M$ poses and substituting them into~\ref{eq:eye_in_hand_2}, a system of equations is obtained. This system is then solved using a least-squares method to estimate the unknown transformation $X \in SE(3)$ \cite{Tsai1988ANT, Jiang2021AnOO, Horn1987ClosedformSO}.  Once $X$ is determined, given the transformation between the robot world frame $\{W\}$ and $\{EE\}$ from the robot’s forward kinematics, the camera scans can be accurately transformed into the world frame $\{W\}$.

\begin{figure}[]
   \centering
   \includegraphics[width=1.0\linewidth]{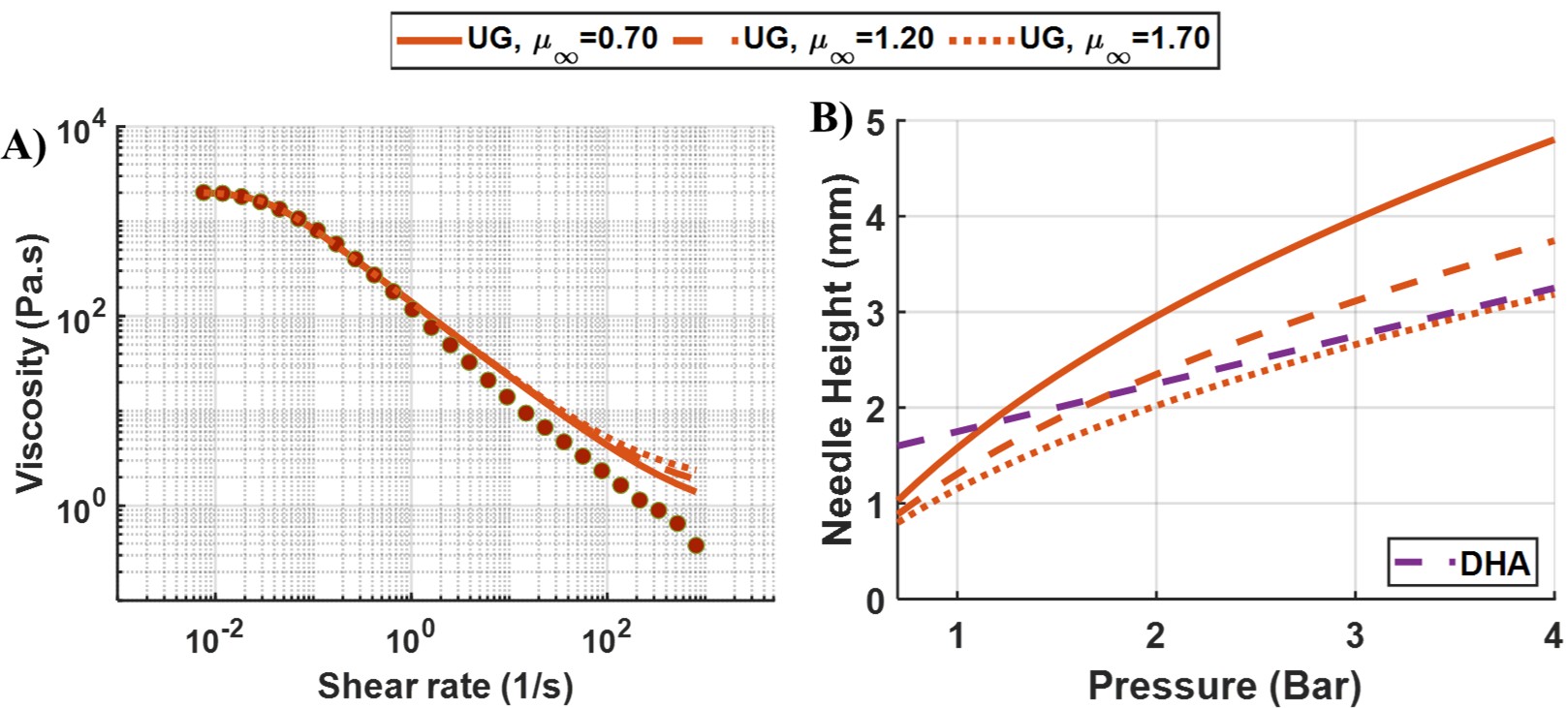}
   \vspace{-6mm}
   \caption{Material characterization. A) Plot shows the results of the rheology tests (solid dots) on the printing material (i.e., ultrasound gel), along with three fitted models using the Carreau model. B) The parameters of the fitted Carreau models are used to simulate the ideal needle height. The DHA model used in this study is also shown. }
   \label{fig:NH}
   \vspace{-3mm}
\end{figure}

\begin{figure*}[]
   \centering
   \includegraphics[width=1.0\linewidth]{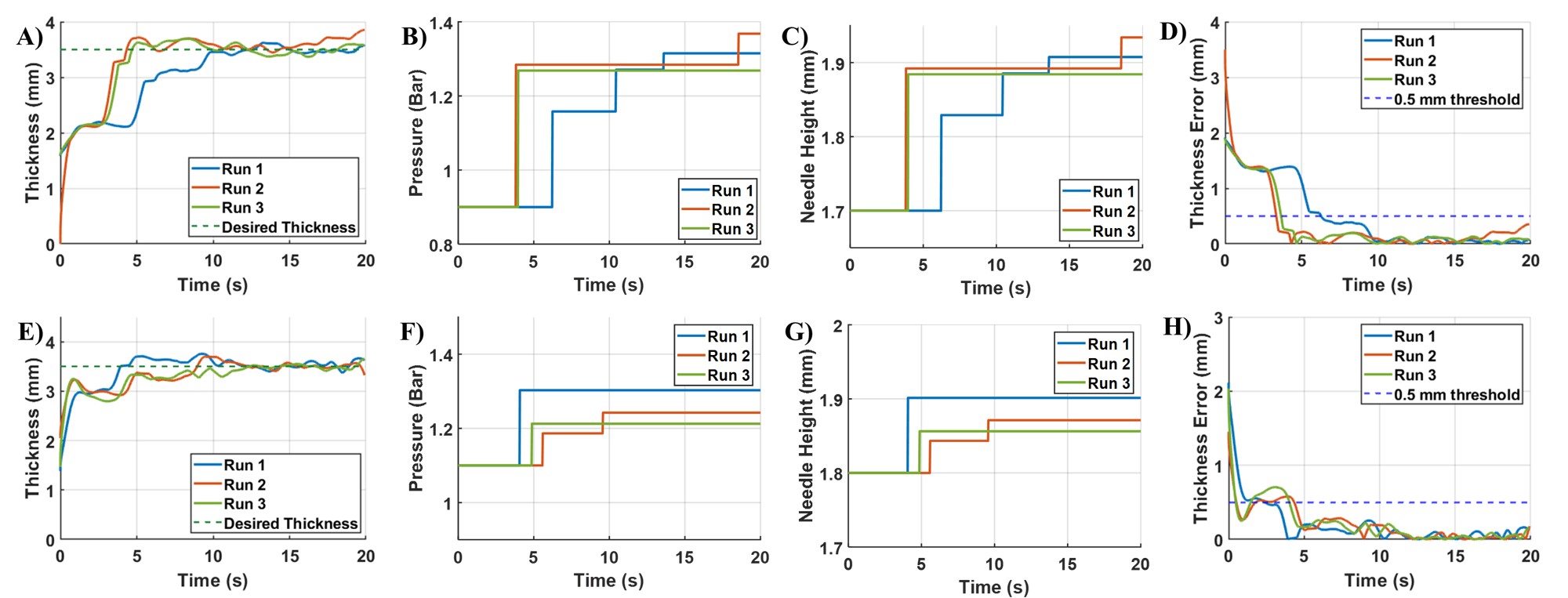}
   \caption{Results of test configurations 1 (A-D) and 2 (E-H). The plots show (A) the measured and desired thickness values, (B) controlled printing pressure $P$, (C) controlled needle height $H$, and (D) the thickness error (i.e., absolute difference between desired and actual thickness) over the time of printing for three repetitions.}
   \label{fig:res_1}
\end{figure*}

\subsection{Delineation and Thickness Calculation Algorithms} \label{subsec:cv_algs}

In this paper, we use a vision-based method for calculating bead thicknesses intermittently. The bead thickness measurement is used to provide feedback to the framework and updating the printing parameters on the fly \cite{ZobeidiAut}. 
This evaluation approach offers significant advantage over commonly used manual measurement methods or other existing ML-based methods for evaluating bead thickness, as it delivers autonomous and robust performance without any need for prior training, making it practical and efficient in printing and clinical applications. 
Furthermore, studies have previously explored offline bead thickness calculation algorithms in DIW bioprinting, as a way of quantitative measurement of the bead geometry \cite{Yang2024DevelopmentAQ, Rezayof2024QuantitativeEO}. 
In these approaches, upon completion of the print, the 3D camera is used to scan and measure the thickness of the beads.
However, these proposed methods are not suitable for online measurement and closed-loop control, due to issues such as reliance on user inputs, slow processing time, and inability to address nozzle masking the beads. 
Thus, here, we use a novel method for calculating the bead thickness, which is well-suited for near real-time calculation and closed-loop control due to its fast processing time, efficiency and complete autonomy.  
The vision-based thickness calculation process (as shown in Fig. \ref{fig:thickness}) includes 6 steps:

\subsubsection{Step I - Scanning} The camera is used to capture a scan from the print site. (See Fig.~\ref{fig:thickness}A).
The printing needle's tip is within the scan, and will be masked in later steps.

\begin{figure*}[]
   \centering
   \includegraphics[width=1.0\linewidth]{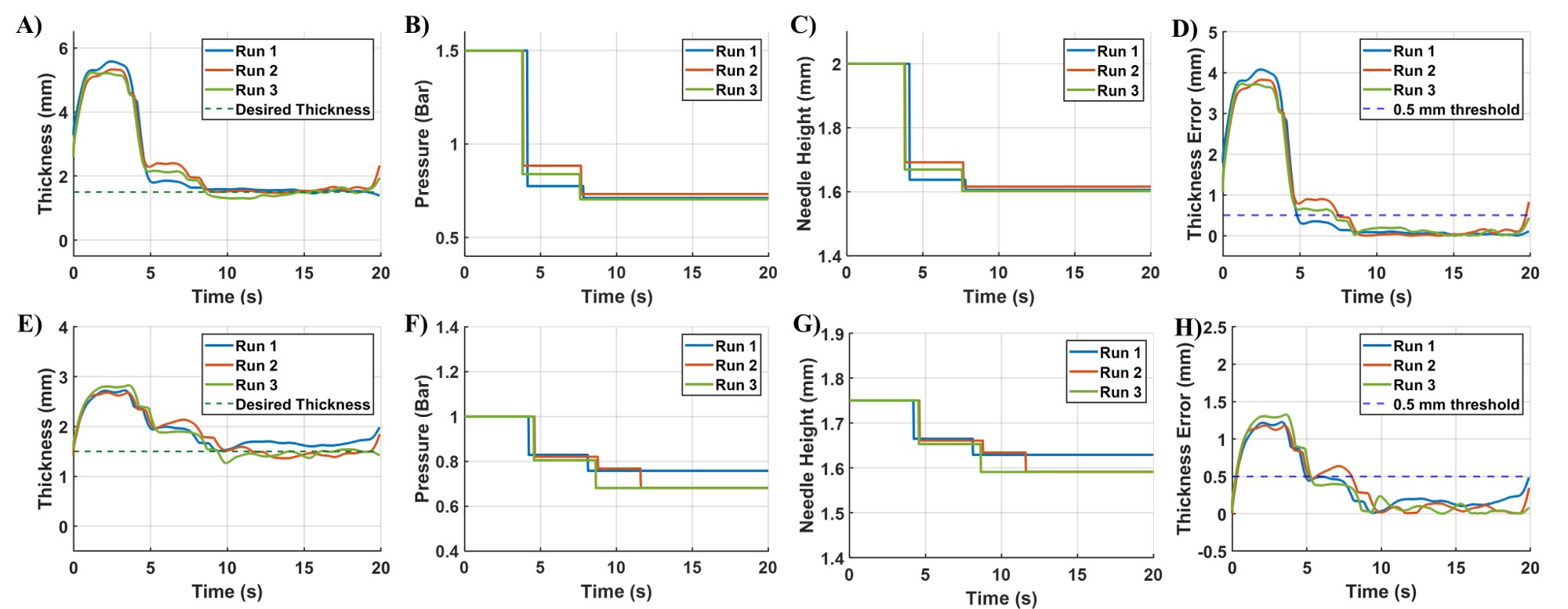}
   \caption{Results of test configurations 3 (A-D) and 4 (E-H). The plots show (A) the measured and desired thickness values, (B) controlled printing pressure $P$, (C) controlled needle height $H$, and (D) the thickness error (i.e., absolute difference between desired and actual thickness) over the time of printing for three repetitions.}
   \label{fig:res_2}
\end{figure*}

\subsubsection{Step II - Grayscale conversion and mask definition}
The scan is converted to grayscale to reduce the processing time by combining three channels of data (red, green and blue) into one channel (gray). 
Furthermore, a polygon is fitted to the bioprinter to allow masking the tip from the Region of Interest (RoI) in the next steps. 

\subsubsection{Step III - Masking}
The polygon is rendered dynamically based on the lighting conditions to make the masked polygon match the printing background. 
In this method, the grayscale value of each pixel row within the polygon mask is replaced with the mean grayscale value of the pixels lying outside the polygon on the same row. This produces a seamless mask of the bioprinter tip within the RoI (see Fig.~\ref{fig:thickness}C).

\subsubsection{Step IV - Edge detection}
Canny edge detector from OpenCV is used to perform edge detection. 
The edge detector provides pixel contours in regions where the image intensity gradient exceeds a threshold. As illustrated in Fig.~\ref{fig:thickness}D, the Canny edge detection result contains discontinuous contours, which requires further refinement.

\subsubsection{Step V - Contour Tracing} \label{subsubsec:delineation} 
In this step, separate contours are merged into unified shapes, and any inner contours detected within the beads are identified and removed to obtain a cleaner representation.
Then, the robot's printing path is overlaid on top of the segmented contours. 

\subsubsection{Step VI - Thickness calculation} \label{subsubsec:thickness_calc}
After bead segmentation, the thickness is calculated as the Euclidean distance between the two boundary points on the bead's borders corresponding to each point in the printing path (Fig. \ref{fig:thickness}F).

\begin{figure*}[]
   \centering
   \includegraphics[width=1.0\linewidth]{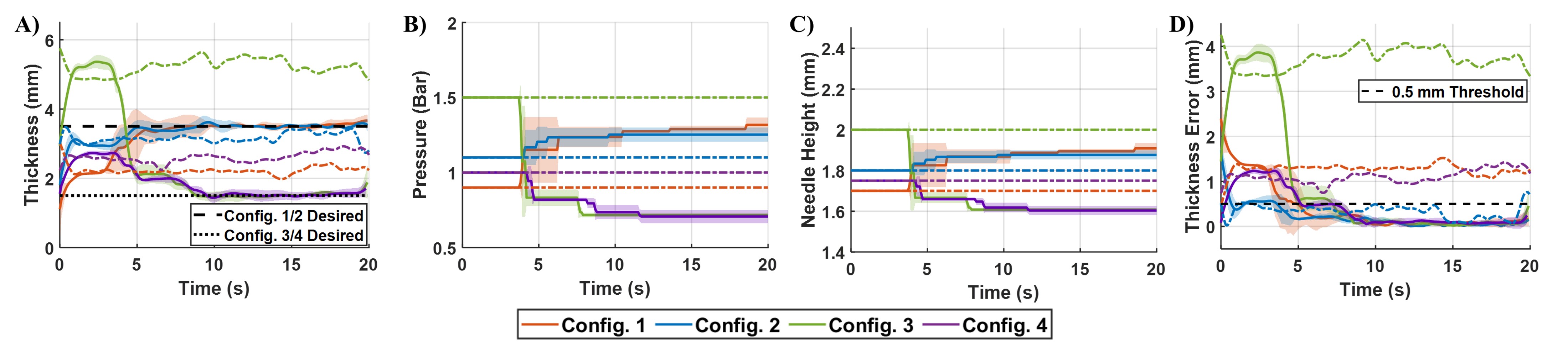}
   \caption{Aggregated results, showing the mean (solid lines) and standard deviation (shaded regions) over the three trials for each test configuration. The figure features the (A) actual thickness and the desired thickness values for each test case, (B) controlled input pressure $P$, (C) controlled needle height $H$, and (D) the thickness error (i.e., absolute difference between the desired and actual thickness values) and a 0.5~mm threshold. The figure further shows the results from open-loop tests are shown in dash-dotted lines with respective colors for each test case.}
   \label{fig:res_aggr}
\end{figure*}

\section{Results \& Discussion} \label{sec:results_and_discussion}
To validate the efficacy of the proposed framework, several experimental studies have been conducted. In these experiments, a target (i.e., desired) thickness was set, and the print would start with an arbitrary pressure value. As the initial pressure is arbitrary, the thickness of the extrudate would be different than the target thickness. Thus, the controller would step in and change the pressure accordingly to track the desired thickness value. 

The experiments are performed using an ultrasound gel (Absonic Conductive Gel, Absonic) dyed in blue. To determine the parameters of the DHA function, a sample of the ultrasound gel was subjected to typical rheology tests as seen in Fig.~\ref{fig:NH}A, to find the viscosity-shear rate relationship. Of note, this test is not material intensive and is usually performed in the development stages of designing a new bioink.
The parameters of the fitted Carreau models are shown in Table~\ref{tab:NH_results}. As explained in Section~\ref{subsec:h_controller}, to reduce time complexity, the controller uses a reduced-order model for controlling the needle height (DHA) instead of relying on the complex iterative-solving methods based on the Carreau model. In this study, $h_0=1.25$ mm and $\alpha=0.5 \text{mm}/\text{bar}$ were used as the DHA parameters for the evaluation experiments. It should be mentioned that, using the DHA function, no occurrences of needle-to-bead penetration happened in the experiments.

The results of these experiments are shown in Fig.~\ref{fig:res_1} and Fig.~\ref{fig:res_2} and Table~\ref{tab:results}. The figures depict four different plots, including the measured thickness, controlled pressure, controlled needle height, and the thickness error over time for different test configurations. Each test configuration was repeated three times to ensure repeatability. As shown in Fig.~\ref{fig:res_1}~and~\ref{fig:res_2} and Table~\ref{tab:results}, the tests were performed with 2 different target thickness values, each with 2 different initial pressures, resulting in a total of 4 combinations of printing parameters. The robot velocity was set to  6 $\text{mm}/\text{s}$ in all tests. Each trial included printing a line of length 120~mm.  
The integral gain ${K_i}$ was selected using the stability limit tuning method. In this method, the integral gain is gradually increased (here in steps of 0.05) until the system becomes unstable. 
The highest value before instability occurs is called $K_{i, ultimate}$ and was found to be 0.4. Following common practice such as in Ziegler-Nichols \cite{ZN}, the controller gain was chosen to be $K_i=0.5K_{i, ultimate}=0.2$ to prevent possible instabilities. 

Table~\ref{tab:results} further shows the details and results from these 4 test configurations. 
The convergence time, as included in the table, represents the total time taken for the system to reach convergence. 
In this study, convergence is defined when the thickness error drops and stays below 0.5 mm. 
As shown in the table, average convergence times were ranging from 3.6 to 6.5 seconds based on the combination of the initial pressure and desired thicknesses. 
Table~\ref{tab:results} provides further insight into the framework's timing performance. The table reports the average consumed time by the framework to (i) perform the scan (limited by the camera performance), and (ii) process the scanned point cloud and RGB images to obtain the thickness values. 
 To further show the advantage of the proposed closed-loop control framework, same printing tests were then repeated under identical conditions, but with the controller disabled. This provides a comparison between the proposed closed-loop framework and traditional open-loop DIW process.
The results of these open-loop tests are shown in Fig.~\ref{fig:res_aggr} as dash-dotted lines.

As Fig.~\ref{fig:NH}A shows, the viscosity of the printing gel significantly drops under higher shear rates, showing a typical shear-thinning non-Newtonian behavior for bioinks ~\cite{Propst2025TimeCF, Kamaraj2024DevelopmentOS}. Figure~\ref{fig:NH} and Table~\ref{tab:NH_results} further prove the limitations of pure and open-loop model-based control of the bead thicknesses, using fluid dynamics modeling. As presented in the table, three different models with different parameters have been fitted to the experimental data, and used to simulate the ideal needle height in Fig.~\ref{fig:NH}B. While all these models are equally good in representing the results of the rheology tests, they result in very different values for the ideal needle heights. This further highlights why pure fluid dynamics-based modeling is insufficient in realistic experiments and the superiority of online feedback-based control in parameter adjusting for desired bead shapes. 

Results show that the developed robotic bioprinting framework was able to successfully track desired values of bead thickness by controlling the pressure and the needle height. Looking at these results in Fig.~\ref{fig:res_1} and Fig.~\ref{fig:res_2}, it can be seen that the controller's performance has been consistent over the three trials for all the Tests. For instance, in configurations 1 and 2, the controller has increased the pressure (see Fig.~\ref{fig:res_1}B and E) to reduce the error (under-deposition), whereas in configurations 3 and 4, the controller has decreased the pressure effectively to reduce the material over-deposition (see Fig.~\ref{fig:res_2}B and E). 
While ML-based methods typically rely on uninterpretable features and can only provide a binary classification of over-/under-deposition, \cite{Bonatti2022ADL, Kelly2025}, the proposed autonomous thickness calculation method directly measures and controls the bead thickness in near real time, enabling precise and smooth control.
As reported in the table, average processing time (0.38-0.45 seconds) is considerably lower than the average time consumed by the camera (1.23-1.68 seconds) showing the time-efficiency of the algorithms used and the potential for future improvement by using newer generations of the point cloud camera. 

Figure~\ref{fig:res_aggr} shows the mean and the standard deviation over the three trials for each case. 
As the figure shows, all test configurations have successfully converged to the desired thickness values. 
Figure~\ref{fig:res_aggr} further demonstrates the advantage and effectiveness of the developed framework. While the thickness error values have remained as high as 5 mm in the open-loop tests, all error values were dropped below 0.5 mm when the controller was on.
Furthermore, Fig.~\ref{fig:res_aggr} and Table~\ref{tab:results} provide insight into the repeatability of the proposed framework. This is shown in the small shaded region over the actual thickness values in the plots. Also, the maximum value for the standard deviation of the converged pressure over all test configurations was found to be 0.05 bar (see Table~\ref{tab:results}). Such small variations over different tests show the repeatability and robustness of the framework. 

Finally, a video demonstration of a case study is provided as a supplementary video. 
In this case study, the print started with a pre-identified pressure corresponding to a specific thickness value. The system was subjected to a change in the desired thickness (step disturbance) during the print. The video shows the system's response to this input disturbance and the successful control of the printing pressure and the needle height to reach and track the new desired thickness value.   
All together, the results show the effectiveness of the framework in controlling the printing parameters (i.e., pressure and needle height) using a vision-based feedback-driven and closed-loop control scheme, to track desired bead thickness values with minimal material-dependent parameter tuning.   

\begin{table}
	\centering
    
	\caption{Fitted Carreau Models' Parameters}
	\label{tab:NH_results}
	
	\begin{tabular}{|c|c|c|c|c|}
		\hline
		  $\mu_{\infty}$ (Pa.s) & $\mu_0$ (Pa.s) & $n$ & $\lambda$ & $R^2$ \\
		\hline
		0.7 & 2044.2296 & 0.2090 & 29.7822 & 0.9995 \\
		1.2 & 2044.0475 & 0.2078 & 29.7422 & 0.9995 \\
		  1.7 & 2043.8660 & 0.2066 & 29.7023 & 0.9995 \\
		
		\hline
	\end{tabular}
\end{table}

\setlength{\tabcolsep}{3.2pt}

\begin{table}
	\centering
    
	\caption{Results of the 4 configurations (standard deviation included in some entries)}
	\label{tab:results}
	
	\begin{tabular}{|c|cccc|}
		\hline
		Parameter & Config. 1 & Config. 2 & Config. 3 & Config. 4 \\
		\hline
		Desired $d$ (mm) 
		    & 3.5 & 3.5 & 1.5 & 1.5 \\
		Initial Pres. (bar) 
		    & 0.9 & 1.1 & 1.5 & 1.0 \\
		Convergence Time (s) 
		    & 4.44$\pm$1.61 & 3.60$\pm$1.05 & 6.54$\pm$1.48 & 6.08$\pm$1.69 \\
		Converged Pres. (bar) 
		    & 1.32$\pm$0.05 & 1.25$\pm$0.05 & 0.72$\pm$0.01 & 0.71$\pm$0.04 \\
		Capture Time (s) 
		    & 1.29$\pm$0.17 & 1.68$\pm$0.30 & 1.25$\pm$0.17 & 1.23$\pm$0.14 \\
		Processing Time (s) 
		    & 0.44$\pm$0.19 & 0.38$\pm$0.13 & 0.44$\pm$0.16 & 0.45$\pm$0.26  \\
		\hline
	\end{tabular}
\end{table}

\section{Conclusion and Future Work}\label{sec:conclusion}
In this paper, we proposed a physics-based closed-loop robotic bioprinting framework for adaptive control of the printing parameters to achieve desired geometrical features (thickness) of the printed constructs. 
The method replaces material and time demanding design of experiments methods for selecting printing parameters by leveraging a closed-loop control system using vision-based feedback. 
Novel algorithms are proposed which enable, for the first time, fully autonomous and near real-time quantitative evaluation of the printed constructs. This evaluation is fed into the control algorithms to further control the printing pressure based on the desired value of the thickness and the feedback. 
Four sets of experiments were designed and conducted to include different scenarios (increasing or decreasing thickness) with two desired thickness values of 1.5 and 3.5 mm, and a range of initial pressure from 0.9 to 1.5 bar. The framework was able to converge to the desired thickness values in all cases, in 5.2 seconds on average.
Moreover, similar experiments were performed using identical printing configuration, but with the controller module turned off, turning the framework into a typical open-loop DIW setup. 
These experiments proved the robustness and effectiveness of the system.  As mentioned before, the framework's capabilities were held back by the camera's capturing time (1.36 seconds on average) which was nearly 3 times the time consumed by the processing algorithms (0.43 seconds on average). 
Thus, hardware improvements in the future work can even further enhance the performance. 
Future work will also include printing using biomaterials specifically designed for VML treatment (e.g.~\cite{Kamaraj2024DevelopmentOS}). 

%\addtolength{\textheight}{-7cm}   % This command serves to balance the column lengths
                                  % on the last page of the document manually. It shortens
                                  % the textheight of the last page by a suitable amount.
                                  % This command does not take effect until the next page
                                  % so it should come on the page before the last. Make
                                  % sure that you do not shorten the textheight too much.

%%%%%%%%%%%%%%%%%%%%%%%%%%%%%%%%%%%%%%%%%%%%%%%%%%%%%%%%%%%%%%%%%%%%%%%%%%%%%%%%

%%%%%%%%%%%%%%%%%%%%%%%%%%%%%%%%%%%%%%%%%%%%%%%%%%%%%%%%%%%%%%%%%%%%%%%%%%%%%%%%

%%%%%%%%%%%%%%%%%%%%%%%%%%%%%%%%%%%%%%%%%%%%%%%%%%%%%%%%%%%%%%%%%%%%%%%%%%%%%%%%

%%%%%%%%%%%%%%%%%%%%%%%%%%%%%%%%%%%%%%%%%%%%%%%%%%%%%%%%%%%%%%%%%%%%%%%%%%%%%%%%

\bibliographystyle{IEEEtran}
\bibliography{main}

\end{document}